\documentclass[runningheads]{llncs}
\usepackage{cvprabbreviation}
\usepackage{graphicx}
\usepackage{comment}
\usepackage{amsmath,amssymb} % define this before the line numbering.
\usepackage{color}

\usepackage{caption}
\usepackage{subfigure}
\usepackage{lipsum}
\usepackage{array}
\usepackage{bbding}
\usepackage{multirow}
\usepackage{cite}
\usepackage[american]{babel}
\usepackage[breaklinks=true,letterpaper=true,colorlinks,bookmarks=false]{hyperref}

\begin{document}
% \linenumbers
\pagestyle{headings}
\mainmatter
\def\ECCVSubNumber{1647}  % Insert your submission number here

\title{Learning Flow-based Feature Warping for \\ Face Frontalization with \\ Illumination Inconsistent Supervision} % Replace with your title

\begin{comment}
\titlerunning{ECCV-20 submission ID \ECCVSubNumber}
\authorrunning{ECCV-20 submission ID \ECCVSubNumber}
\author{Anonymous ECCV submission}
\institute{Paper ID \ECCVSubNumber}
\end{comment}
%******************

%\begin{comment}
\titlerunning{Flow-based Feature Warping Model}

\author{Yuxiang Wei\inst{1} \and Ming Liu\inst{1} \and  Haolin Wang\inst{1} \and Ruifeng Zhu\inst{2, 4} \and  Guosheng Hu \inst{3} \and Wangmeng Zuo\inst{1, 5}$^{(}$\Envelope$^)$} 

\authorrunning{Y. Wei \etal.}
\institute{\textsuperscript{1} School of Computer Science and Technology, Harbin Institute of Technology, China \\
\textsuperscript{2} University of Burgundy Franche-Comté, France  \quad \textsuperscript{3} Anyvision, UK \\
\textsuperscript{4} University of the Basque Country, Spanish \quad  \textsuperscript{5} Peng Cheng Lab, China \\
\email{yuxiang.wei.cs@gmail.com, \{csmliu, Why\_cs\}@outlook.com} \\ 
\email{\{reefing.z, huguosheng100\}@gmail.com, wmzuo@hit.edu.cn} }
%\and
%Peng Cheng Lab, China}
%\end{comment}
%******************align the profile features to frontal view to 
\maketitle

\begin{abstract}
	
Despite recent advances in deep learning-based face frontalization methods, photo-realistic and illumination preserving frontal face synthesis is still challenging due to large pose and illumination discrepancy during training. We propose a novel Flow-based Feature Warping Model (FFWM) which can learn to synthesize photo-realistic and illumination preserving frontal images with illumination inconsistent supervision. Specifically, an Illumination Preserving Module (IPM) is proposed to learn illumination preserving image synthesis from illumination inconsistent image pairs. IPM includes two pathways which collaborate to ensure the synthesized frontal images are illumination preserving and with fine details. Moreover, a Warp Attention Module (WAM) is introduced to reduce the pose discrepancy in the feature level, and hence to synthesize frontal images more effectively and preserve more details of profile images. The attention mechanism in WAM helps reduce the artifacts caused by the displacements between the profile and the frontal images. Quantitative and qualitative experimental results show that our FFWM can synthesize photo-realistic and illumination preserving frontal images and performs favorably against the state-of-the-art results. Our code is available at \url{https://github.com/csyxwei/FFWM}.


\end{abstract}

\section{Introduction}

Face frontalization aims to synthesize the frontal view face from a given profile. Frontalized faces can be directly used for general face recognition methods without elaborating additional complex modules. Apart from face recognition, generating photo-realistic frontal face is beneficial for a series of face-related tasks, including face reconstruction, face attribute analysis, facial animation, \etc.

Traditional methods address this problem through 2D/3D local texture warping \cite{hassner2015effective, zhu2015high} or statistical modeling\cite{sagonas2015robust}. Recently, GAN-based methods have been proposed to recover a frontal face in a data-driven manner \cite{huang2017beyond, zhao2018towards, tran2017disentangled, yin2017towards, hu2018pose, zhao20183d, cao2018learning, yin2020dual}. For instance, Yin \etal \cite{yin2020dual} propose DA-GAN to capture the long-displacement contextual information from illumination discrepancy images under large poses. However, it recovers inconsistent illumination on the synthesized image. Flow-based method \cite{zhang2018face} predicts a dense pixel correspondence between the profile and frontal image and uses it to deform the profile face to the frontal view. However, deforming the profile face in the image space directly leads to obvious artifacts and missing pixels should be addressed under large poses.

\begin{figure}[t]
	\centering
	\includegraphics[width=0.85\textwidth]{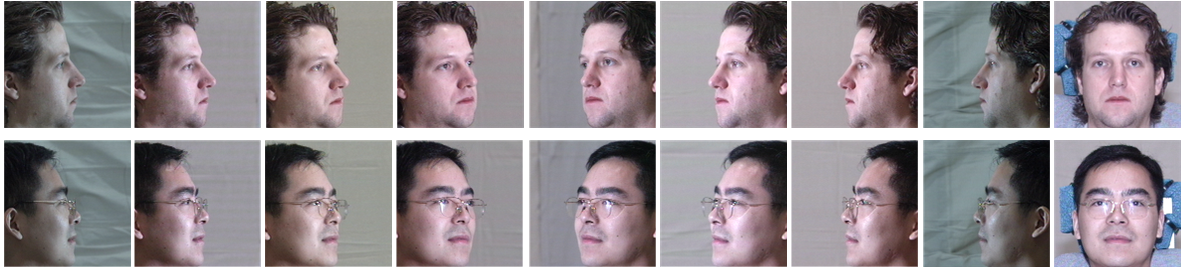}
	\caption{$ \pm 45^\circ $, $ \pm 60^\circ $, $ \pm 75^\circ $ and $ \pm 90^\circ $ images of the two persons in the Multi-PIE. Each row images have the same flash in the recording environment.}
	\label{fig:iilu_inconsist}
\end{figure}

The existing methods do not consider the illumination inconsistency between the profile and ground-truth frontal image. Taking the widely used benchmark Multi-PIE \cite{gross2010multi} as an example, the visual illumination conditions on several poses are significantly different from the ground-truth frontal images as shown in Fig. \ref{fig:iilu_inconsist}. Except $ \pm 90^\circ $, the other face images are produced by the same camera type. The variation in camera types causes obvious illumination inconsistency between the $ \pm 90^\circ $ images and the ground-truth frontal image. Although efforts have been made to manually color-balance those same type cameras, the illumination of resulting images within  $ \pm 75^\circ $ (except $ 0^\circ $) still look visually distinguishable with the ground-truth frontal image. Since the existing methods minimize pixel-wise loss between the synthesized image and the illumination inconsistent ground-truth, they tend to change both the pose and the illumination of the profile face image, while the latter actually is not acceptable in face editing and synthesis.

To address the above issue, this paper proposes a novel Flow-based Feature Warping Model (FFWM) which can synthesize photo-realistic and illumination preserving frontal image from illumination inconsistent image pairs. In particular, FFWM incorporates the flow estimation with two modules: Illumination Preserving Module (IPM) and Warp Attention Module (WAM). Specifically, we estimate the optical flow fields from the given profile: the reverse and forward flow fields are predicted to warp the front face to the profile view and vice versa, respectively. The estimated flow fields are fed to IPM and WAM to conduct face frontalization.

The IPM is proposed to synthesize illumination preserving images with fine facial details from illumination inconsistent image pairs. Specifically, IPM contains two pathways: (1) Illumination Preserving Pathway and (2) Illumination Adaption Pathway. For (1), an illumination preserving loss equipped with the reverse flow field is introduced to constrain the illumination consistency between synthesized images and the profiles. For (2), guided filter \cite{he2010guided} is introduced to further eliminate the illumination discrepancy and learns frontal view facial details from illumination inconsistent ground-truth image. The WAM is introduced to reduce the pose discrepancy in the feature level. It uses the forward flow field to align the profile features to the frontal view. This flow provides an explicit and accurate supervision signal to guide the frontalization. The attention mechanism in WAM helps to reduce the artifacts caused by the displacements between the profile and frontal images.

Quantitative and qualitative experimental results demonstrate the effectiveness of our FFWM on synthesizing photo-realistic and illumination preserving faces with large poses and the superiority over the state-of-the-art results on the testing benchmarks. Our contributions can be summarized as:

\begin{itemize}
	\item A Flow-based Feature Warping Model (FFWM) is proposed to address the challenging problem in face frontalization, \ie photo-realistic and illumination preserving image synthesis. 
	\item Illumination Preserving Module (IPM) equipped with guided filter and flow field is proposed to achieve illumination preserving image synthesis. Warp Attention Module (WAM) uses the attention mechanism to effectively reduce the pose discrepancy in the feature level under the explicit and effective guidance from flow estimation.
	\item Quantitative and qualitative results demonstrate that the proposed FFWM outperforms the state-of-the-art methods.
\end{itemize}

\section{Related Work}

\subsection{Face Frontalization}

Face frontalization aims to synthesize the frontal face from a given profile. Traditional methods address this problem through 2D/3D local texture warping \cite{hassner2015effective, zhu2015high} or statistical modeling\cite{sagonas2015robust}. Hassner \etal \cite{hassner2015effective} employ a mean 3D model for face normalization. A statistical model\cite{sagonas2015robust} is used for frontalization and landmark detection by solving a constrained low-rank minimization problem.

Benefiting from deep learning, many GAN-based methods\cite{huang2017beyond,tran2017disentangled,zhang2018face,hu2018pose,tian2018cr, yin2020dual} are proposed for face frontalization. Huang \etal \cite{huang2017beyond} use a two-pathway GAN architecture for perceiving global structures and local details simultaneously. Domain knowledge such as symmetry and identity information of face is used to make the synthesized faces photo-realistic. Zhao \etal \cite{zhao2018towards} propose PIM with introducing a domain adaptation strategy for pose invariant face recognition.
3D-based methods \cite{yin2017towards,zhao20183d,deng2018uv,cao2018learning} attempt to combine prior knowledge of 3D face with face frontalization. Yin \etal \cite{yin2017towards} incorporate 3D face model into GAN to solve the problem of large pose face frontalization in the wild. HF-PIM \cite{cao2018learning} combines the advantages of 3D and GAN-based methods and frontalizes profile images via a novel texture warping procedure. In addition to supervised learning, Qian \etal \cite{qian2019unsupervised} propose a novel Face Normalization Model (FNM) for unsupervised face generation with unpaired face images in the wild. Note that FNM focuses on face normalization, without considering preserving illumination.

Instead of learning function to represent the frontalization procedure, our method gets frontal warped feature by flow field and reconstructs illumination preserving and identity preserving frontal view face.

\subsection{Optical Flow}

Optical flow estimation has many applications, \eg, action recognition, autonomous driving and video editing. With the progress in deep learning, FlowNet\cite{dosovitskiy2015flownet}, FlowNet2\cite{ilg2017flownet} and others achieve good results by end-to-end supervised learning. While SpyNet\cite{ranjan2017optical}, PWC-Net\cite{sun2018pwc} and LiteFlowNet\cite{hui2018liteflownet} also use coarse-to-fine strategery to refine the initial flow. It is worth mentioning that PWC-Net and LiteFlowNet have smaller size and are easier to train. Based on weight sharing and residual subnetworks, Hur and Roth \cite{hur2019iterative} learn bi-directional optical flow and occlusion estimation jointly. Bilateral refinement of flow and occlusion address blurry estimation, particularly near motion boundaries. By the global and local correlation layers, GLU-Net\cite{truong2019glu} can resolve the challenges of large displacements, pixel-accuracy, and appearance changes.

In this work, we estimate bi-directional flow fields to represent dense pixel correspondence between the profile and frontal faces, which are then exploited to obtain frontal view features and preserve illumination condition, respectively.

\begin{figure}[t]
	\centering
	\includegraphics[width=1\textwidth]{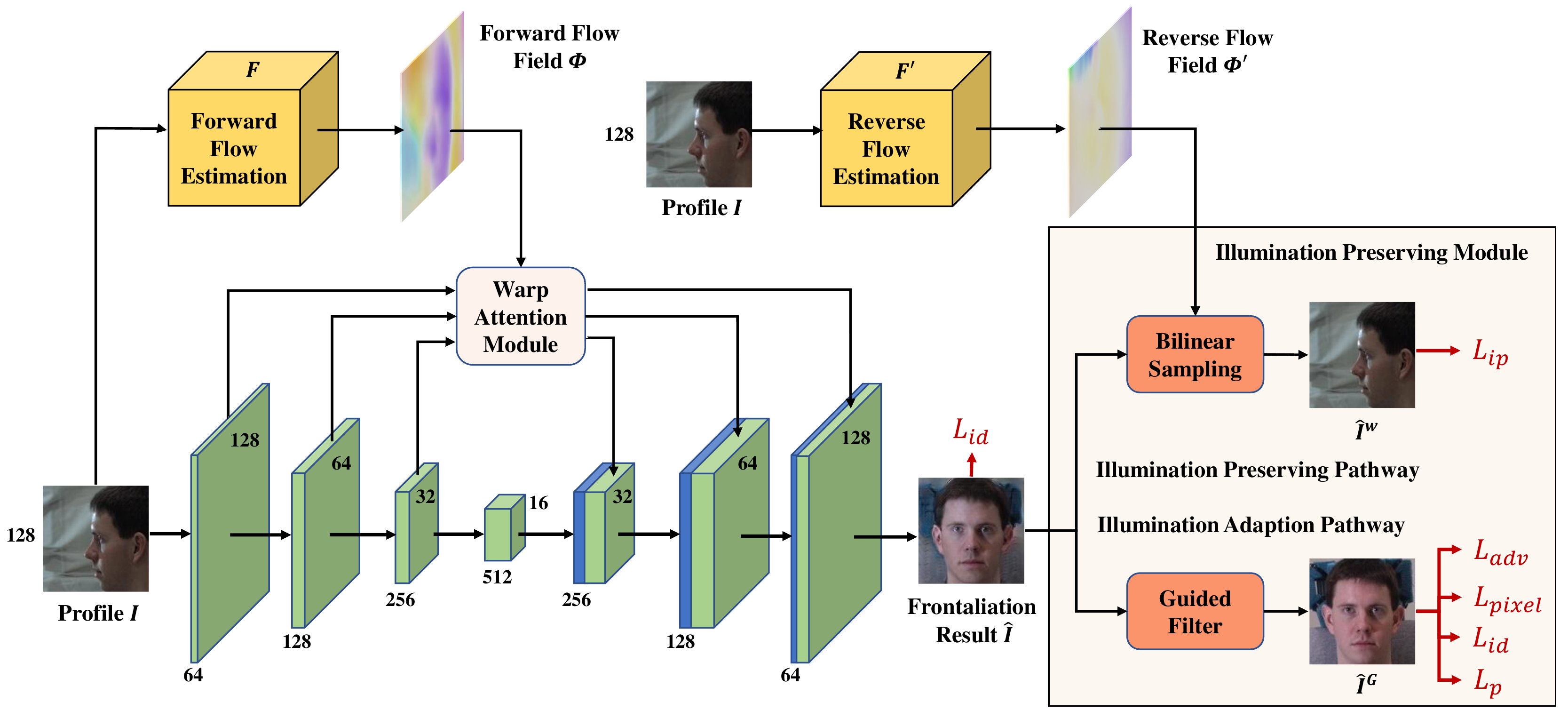}
	\caption{The architecture of our FFWM. Illumination Preserve Module is incorporated to facilitate synthesized frontal image $\hat{I}$ to be illumination preserving and facial details preserving in two independent pathways. Based on the skip connection, the Warp Attention Module helps synthesize frontal image effectively. Losses are shown in red color, which $ \hat{I}^w $ is the synthesized image $ \hat{I} $ warped by $ \Phi' $ and the $ \hat{I}^G $ is the guided filter output. }
	\label{fig:network}
\end{figure}

\section{Proposed Method}

Let \{$I, I^{gt}$\} be a pair of profile and frontal face image of the same person. Given a profile image $ I $, our goal is to train a model $\mathcal{R}$ to synthesize the corresponding frontal face image $ \hat{I} = \mathcal{R}(I) $, which is expected to be photo-realistic and illumination preserving. To achieve this, we propose the Flow-based Feature Warping Model (FFWM). As shown in Fig.~\ref{fig:network}, FFWM takes U-net \cite{ronneberger2015u} as the backbone and incorporates with the  Illumination Preserving Module (IPM) and the Warp Attention Module (WAM) to synthesize $\hat{I}$. In addition, FFWM uses optical flow fields which are fed to IPM and WAM to conduct frontalization. Specifically, we compute the forward and reverse flow fields to warp the profile to the frontal view and vice versa, respectively. 

In this section, we first introduce the bi-direcrional flow fields estimation in Sec~\ref{section:3.1}. IPM and WAM are introduced in Sec~\ref{section:3.2} and Sec~\ref{section:3.3}. Finally, the loss functions are detailed in Sec~\ref{section:3.4}.

\subsection{Bi-directional Flow Fields Estimation \label{section:3.1}}

Face frontalization can be viewed as the face rotation transformation, and the flow field can model this rotation by establishing the pixel-level correspondence between the profile and frontal faces. Traditional optical flow methods \cite{dosovitskiy2015flownet, ilg2017flownet} take two frames as the input. However, we only use one profile image as the input. In this work, we adopt the FlowNetSD in FlowNet2 \cite{ilg2017flownet} as our flow estimation network, and change the input channel from 6 (two frames) to 3 (one image). For preserving illumination and frontalization, we estimate the reverse flow field $ \Phi' $ and the forward flow field $ \Phi $ from the profile image, respectively.

\subsubsection{Reverse Flow Field.}
Given the profile image $ I $, reverse flow estimation network $ \mathcal{F'} $ predicts the reverse flow field $ \Phi' $ which can warp the ground-truth frontal image $ I^{gt} $ to the profile view as $ I $.
\begin{equation}
\Phi' =\mathcal{F'}(I;\Theta_\mathcal{F'}),
\end{equation}
\begin{equation}
{I^w}' = \mathcal{W}(I^{gt},\Phi'),
\end{equation}
where $\Theta_\mathcal{F'}$ denotes the parameters of $ \mathcal{F'} $, and $\mathcal{W}(\cdot)$ \cite{jaderberg2015spatial} is the bilinear sampling operation. To learn an accurate reverse flow field,  $ \mathcal{F}' $ is pretrained with the landmark loss\cite{li2018learning}, sampling correctness loss\cite{ren2020deep} and the regularization term \cite{ren2020deep}.

\subsubsection{Forward Flow Field.} 
Given the profile image $ I $, forward flow estimation network $ \mathcal{F} $ predicts the forward flow field $ \Phi $ which can warp $ I $ to the frontal view.
\begin{equation}
\Phi =\mathcal{F}(I;\Theta_\mathcal{F}),
\end{equation}
\begin{equation}
I^w = \mathcal{W}(I,\Phi),
\end{equation}
where $\Theta_\mathcal{F}$ denotes the parameters of $ \mathcal{F} $. To learn an accurate forward flow field,  $ \mathcal{F} $ is pretrained with the same losses as $ \mathcal{F}' $. 

%and more details can be found in the supplementary material.

Then two flow fields $ \Phi' $ and  $ \Phi $ are used for the IPM and WAM to generate illumination preserving and photo-realistic frontal images.

\subsection{Illumination Preserving Module \label{section:3.2}}

Without considering inconsistent illumination in the face datasets, the existing frontalization methods potentially overfit to the wrong illumination. To effectively decouple the illumination and the facial details, hence to synthesize illumination preserving faces with fine details, we propose the Illumination Preserving Module (IPM).  As shown in Fig. \ref{fig:network}, IPM consists of two pathways. Illumination preserving pathway ensures that the illumination condition of the synthesized image $\hat{I}$ is consistent with the profile $I$. Illumination adaption pathway ensures that the facial details of the synthesized image $\hat{I}$ are consistent with the ground-truth $I^{gt}$.

\subsubsection{Illumination Preserving Pathway.}
Because the illumination condition is diverse and cannot be quantified as a label, it is hard to learn reliable and independent illumination representation from face images. Instead of constraining the illumination consistency between the profile and the synthesized image in the feature space, we directly constrain it in the image space. As shown in Fig. \ref{fig:network}, in the illumination preserving pathway, we firstly use the reverse flow field ${\Phi}'$ to warp the synthesized image $ \hat{I} $ to the profile view,
\begin{equation}
\hat{I}^w =  \mathcal{W}(\hat{I}, \Phi').
\end{equation}
Then an illumination preserving loss is defined on the warped synthesized image $ \hat{I}^w $ to constrain the illumination consistency between the synthesized image $ \hat{I} $ and the profile $ I $. By minimizing it, FFWM can synthesize illumination preserving frontal images.

\subsubsection{Illumination Adaption Pathway.}
Illumination preserving pathway cannot ensure the consistency of facial details between the synthesized image $\hat{I}$ and the ground-truth $ I^{gt} $, so we constrain it in the illumination adaption pathway. Since the illumination of profile $ I $ is inconsistent with the ground-truth $ I^{gt} $ under large poses,  adding constraints directly between $\hat{I}$ and $ I^{gt} $ eliminates the illumination consistency between $ \hat{I}$ and $ I $. So a guided filter layer\cite{he2010guided} is firstly used to transfer the illumination of images. Specifically, the guided filter takes $I^{gt}$ as the input image and $\hat{I}$ as the guidance image,
\begin{equation}
\hat{I}^G = \mathcal{G}(\hat{I},I^{gt}),
\end{equation}
where $ \mathcal{G}(\cdot) $ denotes the guided filter, and we set the radius of filter as the quarter of the image resolution. After filtering, the guided filter result $ \hat{I}^G $ has the same illumination with $ I^{gt} $ while keeping the same facial details with $ \hat{I} $. Then the illumination-related losses (\eg, pixel-wise loss, perceptual loss) are defined on $ \hat{I}^G $ to facilitate our model synthesize $ \hat{I} $ with much finer details. By this means, $\hat{I}$ can become much more similar to $I^{gt}$ in facial details without changing the illumination consistency between $ \hat{I} $ and $ I $. 

Note that the guided filter has no trainable parameters and potentially cause our model trap into local minima during training. So we apply the guided filter after several iterations, providing stable and robust initialization to our model.

\subsection{Warp Attention Module \label{section:3.3}}

\begin{figure}[t]
	\centering
	\includegraphics[width=0.75\textwidth]{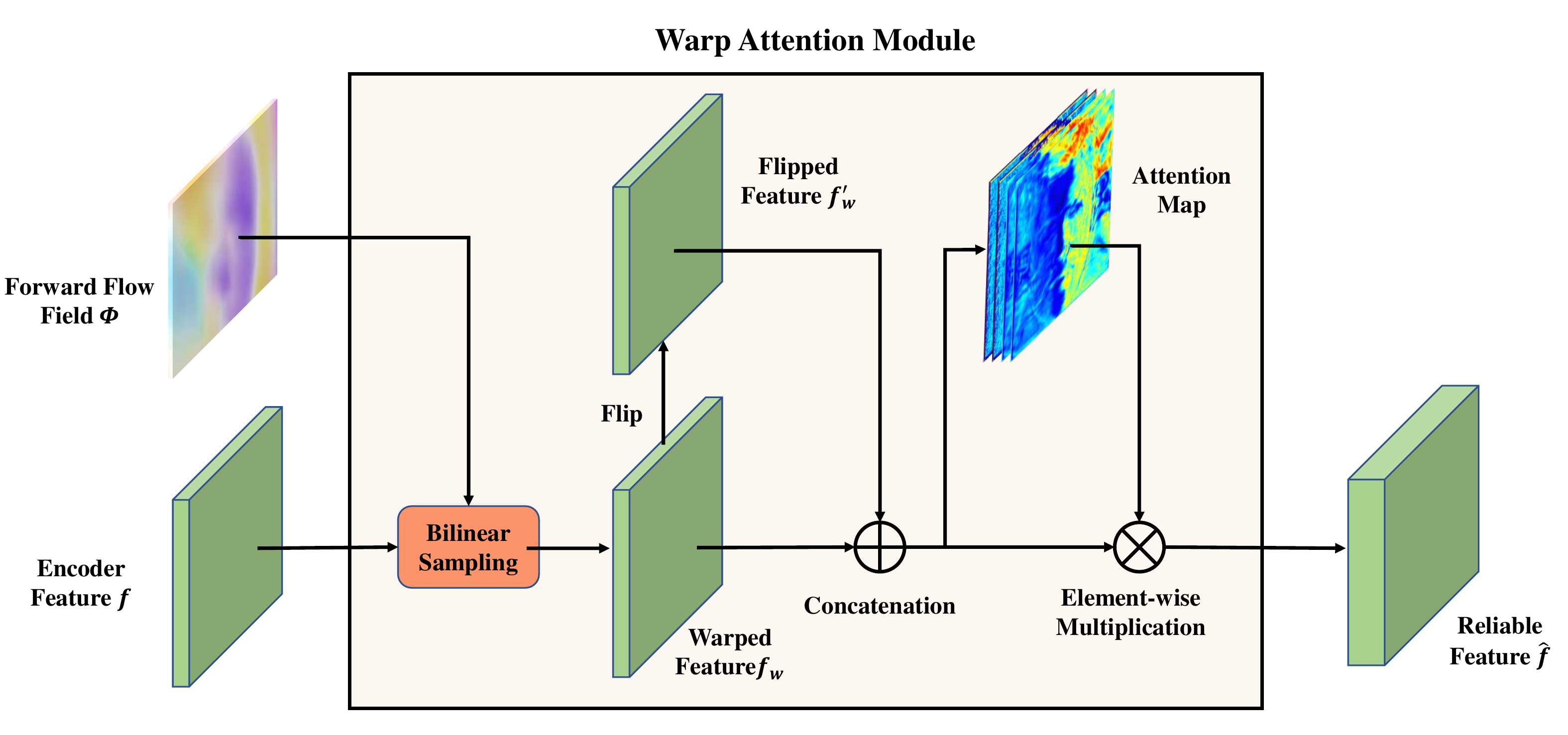}
	\caption{The architecture of Warp Attention Module. Considering the symmetry prior of human face, WAM also contains flipped warped feature.}
	\label{fig:WAM}
\end{figure}

% The synthesized images tend to lost contents due to the large pose discrepancy.
The large pose discrepancy makes it difficult to synthesize correct facial details in the synthesized images. To reduce the pose discrepancy between the profile and frontal face, Warp Attention Module (WAM) is proposed to align the profile face to the frontal one in the feature space. We achieve this by warping the profile features guided by the forward flow field $ \Phi $. The architecture of our WAM is illustrated in Fig. \ref{fig:WAM}. It contains two steps: flow-guided feature warping and feature attention. 

\subsubsection{Flow-Guided Feature Warping.}
Because the profile and frontal face have different visible areas, the forward flow field $ \Phi $ cannot establish a complete pixel-level correspondence between them. Hence, warping profile face directly leads to artifacts. Here we incorporate $ \Phi $ with bilinear sampling operation $\mathcal{W}(\cdot)$ to warp the profile face to the frontal one in the feature space. Additionally, we use the symmetry prior of human face, and take both warped features and its horizontal flip to guide the frontal image synthesis.
\begin{equation}
f_w = \mathcal{W}(f, \Phi) ,
\end{equation}
where $ f $ denotes the encoder feature of the profile. Let $ {f_w}' $ denotes the horizontal flip of $ f_w $, and $ (f_w \oplus {f_w}') $ denotes the concatenation of $ f_w $ and $ {f_w}' $. 

\subsubsection{Feature Attention.}
After warping, the warped feature encodes the backgrounds and self-occlusion artifacts, which leads to degraded frontalization performance. To eliminate above issue and extract reliable frontal feature, an attention mechanism is then used to adaptively focus on the critical parts of $ (f_w \oplus {f_w}') $. The warped feature $ (f_w \oplus {f_w}') $ is firstly fed into a Conv-BN-ReLU-ResidualBlock Layer to generate an attention map $ A $, which has the same height, width and channel size with $ (f_w \oplus {f_w}') $. Then the reliable frontalized feature $\hat{f}$ is obtained by,
\begin{equation}
\hat{f} = A \otimes (f_w \oplus {f_w}') ,
\end{equation}
where $\otimes$ denotes element-wise multiplication. $\hat{f}$ is then skip connected to the decoder to help generate photo-realistic frontal face image $ \hat{I} $.
%Integrating attention mechanism with discriminator produces photo-realistic face components with great quality.

\subsection{Loss Functions \label{section:3.4}}

In this section, we formulate the loss functions used in our work. The background of images is masked to make the loss functions focus on the facial area.  
%To pay more attention to the face synthesis, the background of images participating in loss calculation are masked.

\subsubsection{Pixel-wise Loss.} 
Following \cite{huang2017beyond, hu2018pose}, we employ a multi-scale pixel-wise loss on the guided filter result $ \hat{I}^G $ to constrain the content consistency,
\begin{equation}
\mathcal{L}_{pixel}=  \sum_{s=1}^{S} \left \| \hat{I}^G_s - I^{gt}_s \right \|_1 ,
\label{eq:l1}
\end{equation}
where $ S $ denotes the number of scales. In our experiments, we set $S$ = 3, and the
scales are 32 $\times$ 32, 64 $\times$ 64 and 128 $\times$ 128. 

\subsubsection{Perceptual Loss.} 
%Only using the pixel-wise loss inclines to cause over-smoothing result. 
Pixel-wise loss tends to generate over-smoothing results. To alleviate this, we introduce the perceptual loss defined on the VGG-19 network \cite{simonyan2014very} pre-trained on ImageNet \cite{russakovsky2015imagenet}, 
\begin{equation}
\mathcal{L}_{p}= \sum_{i}w_i \left \| \phi_i(\hat{I}^G) - \phi_i(I^{gt}) \right \|_1 ,
\label{eq:perceptaul}
\end{equation}
where $\phi_i(\cdot)$ denotes the output of the $i$-th VGG-19 layer. In our implementation, we use Conv1-1, Conv2-1, Conv3-1, Conv4-1 and Conv5-1 layer, and set $ w = \{1, 1/2, 1/4, 1/4, 1/8\} $. To improve synthesized imagery in the particular
facial regions, we also use the perceptual loss on the facial regions like eyes, nose and mouth.

\subsubsection{Adversarial Loss.} 
Following \cite{shocher2018ingan}, we adpot a multi-scale discriminator and adversarial learning to help synthesize photo-realistic images.
\begin{equation}
\mathcal{L}_{adv} = \min_{R}\max_{D} \mathbb{E}_{I^{gt}}[ \log D(I^{gt} ) ] - \mathbb{E}_{\hat{I}^G}[\log(1-D(\hat{I}^G))] .
\end{equation}

\subsubsection{Illumination Preserving Loss.} 
To preserve the illumination of profile $ I $ on synthesized image $ \hat{I} $, we define the illumination preserving loss on the warped synthesized image $ \hat{I}^w $ at different scales, 
\begin{equation}
\mathcal{L}_{ip}= \sum_{s=1}^{S} \left \| \hat{I}^w_s - I_s \right \|_1 ,
\end{equation}
where $ S $ denotes the number of scales, and the scale setting is same as Eqn.~(\ref{eq:l1}).

\subsubsection{Identity Preserving Loss.} 

Following \cite{huang2017beyond, hu2018pose}, we present an identity preserving loss to preserve the identity information of the synthesized image $ \hat{I} $,
\begin{equation}
\mathcal{L}_{id}=\left \| \psi_{fc2} (\hat{I})-\psi_{fc2} (I^{gt} )  \right \|_1 + \left \| \psi_{pool } (\hat{I})-\psi_{pool } (I^{gt})  \right \|_1 ,
\end{equation}
where $ \psi(\cdot) $ denotes the pretrained LightCNN-29\cite{wu2018light}. $ \psi_{fc2}(\cdot) $ and $ \psi_{pool}(\cdot) $ denote the outputs of the last pooling layer and the fully connected layer respectively. To preserve the identity information, we add the identity loss on both $ \hat{I} $ and $ \hat{I}^G $.

\subsubsection{Overall Losses.}

Finally, we combine all the above losses to give the overall model objective,
\begin{equation}
\mathcal{L}= \lambda_0 \mathcal{L}_{pixel} + \lambda_1 \mathcal{L}_{p} + \lambda_2 \mathcal{L}_{adv} + \lambda_3 \mathcal{L}_{ip} + \lambda_4 \mathcal{L}_{id},
\end{equation}
where $ \lambda_{*} $ denotes the different losses tradeoff parameters.

\section{Experiments}

To illustrate our model can synthesize photo-realistic and illumination preserving images while preserving identity, we evaluate our model qualitatively and quantitatively under both controlled and in the wild settings. 
In the following subsections, we begin with an introduction of datasets and implementation details. Then we demonstrate the merits of our model on qualitative synthesis results and quantitative recognition results over the state-of-the-art methods. Lastly, we conduct an ablation study to demonstrate the benefits from each part of our model.

\subsection{Experimental Settings}

\begin{figure}[t]
	\centering
	\includegraphics[width=0.95\textwidth]{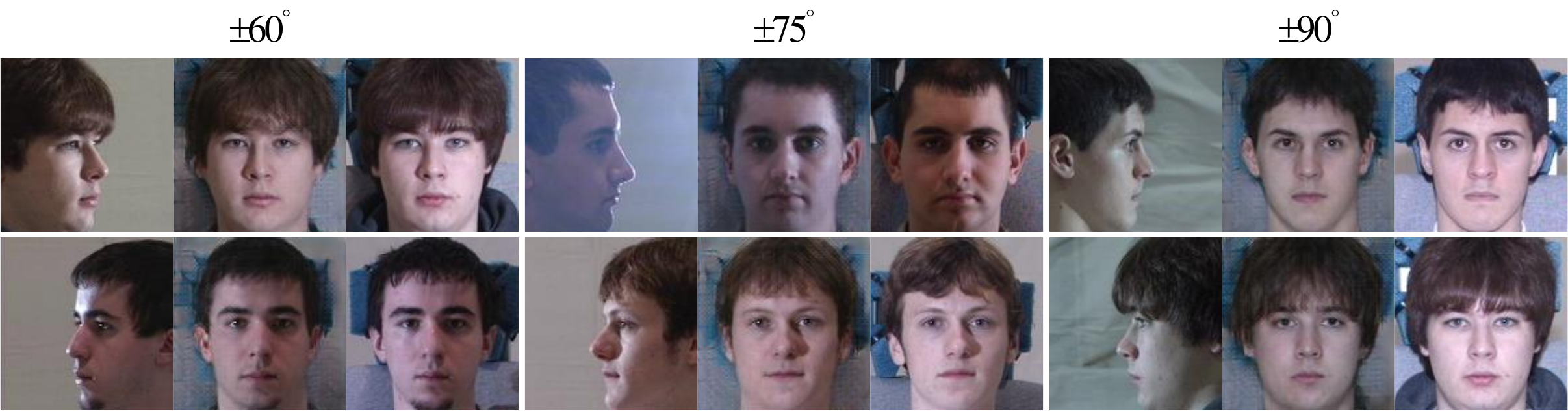}
	\caption{Synthesis results of the Multi-PIE dataset by our model under large poses and illumination inconsistent conditions. Each pair presents profile (left), synthesized frontal face (middile) and ground-truth frontal face (right).}
	\label{fig:illumination}
\end{figure}

\subsubsection{Datasets.} We adopt the Multi-PIE dataset \cite{gross2010multi} as our training and testing set. Multi-PIE is widely used for evaluating face synthesis and recognition in the controlled setting. It contains 754,204 images of 337 identities from 15 poses and 20 illumination conditions. In this paper, the face images with neutral expression under 20 illuminations and 13 poses within $ \pm 90^\circ $  are used. For a fair comparison, we follow the test protocols in \cite{hu2018pose} and utilize two settings to evaluate our model. The first setting (Setting 1) only contains images from Session 1. The training set is composed of all the first 150 identities images. For testing, one gallery image with frontal view and normal illumination is used for the remaining 99 identities. For the second setting (Setting 2), we use neutral expression images from all four sessions. The first 200 identities and the remaining 137 identities are used for training and testing, respectively. Each testing identity has one gallery image with frontal view and normal illumination from the first appearance.

LFW\cite{lfw} contains 13,233 face images collected in unconstrained environment. It will be used to evaluate the frontalization performance in uncontrolled settings.

\subsubsection{Implementation Details.}

All images in our experiments are cropped and resized to 128 $ \times $ 128 according to facial landmarks, and image intensities are linearly scaled to the range of [0, 1]. The LightCNN-29 \cite{wu2018light} is pretrained on MS-Celeb-1M \cite{guo2016ms} and fine-tuned on the training set of Multi-PIE.

In all our experiments, we empirically set $ \lambda_0 = 5, \lambda_1 = 1, \lambda_2 = 0.1, \lambda_3 = 15, \lambda_4 = 1 $. The learning rate is initialized by 0.0004 and the batch size is 8. The flow estimation networks $ \mathcal{F} $ and $ \mathcal{F'} $ are pre-trained and then all networks are end-to-end trained by minimizing the objective $ \mathcal{L} $ with setting lr=0.00005 for $ \mathcal{F} $ and $ \mathcal{F'} $.

\subsection{Qualitative evaluation}

In this subsection, we qualitatively compare the synthesized results of our model against state-of-the-art face frontalization methods. We train our model on the training set of the
Multi-PIE Setting 2, and evaluate it on the testing set of the Multi-PIE Setting 2 and the LFW \cite{lfw}.

Fig. \ref{fig:illumination} shows the face synthesized results under large poses, and it is obvious that our model can synthesize photo-realistic images. To demonstrate the illumination preserving strength of our model, we choose the profiles with obvious inconsistent illumination.
%^having obvious inconsistent illumination compared with the ground-truth images. 
As shown in Fig. \ref{fig:illumination}, the illumination of profile faces can be well preserved in the synthesized images. More synthesized results are provided in the supplementary material.

\begin{figure}[t]
	\centering
	\subfigure[Profile]{
		\begin{minipage}[b]{0.12\linewidth}
			\centering
			\includegraphics[width=0.85\linewidth]{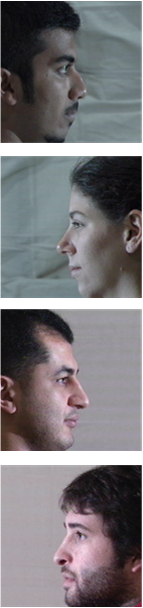}
	\end{minipage}}
	\subfigure[Ours]{
		\begin{minipage}[b]{0.12\linewidth}
			\centering
			\includegraphics[width=0.85\linewidth]{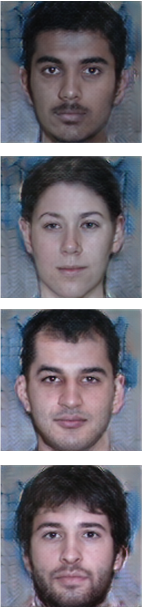}
	\end{minipage}}
	\subfigure[~\cite{hu2018pose}]{
		\begin{minipage}[b]{0.12\linewidth}
			\centering
			\includegraphics[width=0.85\linewidth]{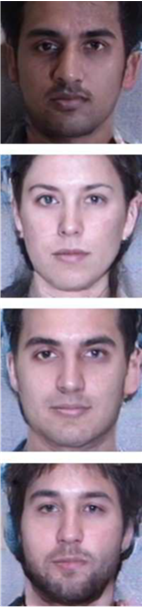}
	\end{minipage}}
	\subfigure[~\cite{huang2017beyond}]{
		\begin{minipage}[b]{0.12\linewidth}
			\centering
			\includegraphics[width=0.85\linewidth]{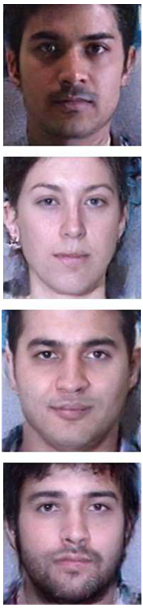}
	\end{minipage}}
	\subfigure[~\cite{yin2017towards}]{
		\begin{minipage}[b]{0.12\linewidth}
			\centering
			\includegraphics[width=0.85\linewidth]{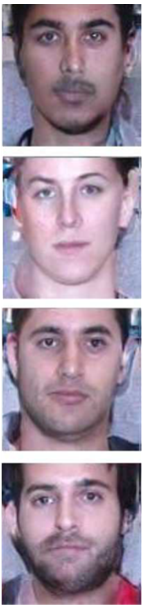}
	\end{minipage}}
	\subfigure[\cite{qian2019unsupervised}]{
		\begin{minipage}[b]{0.12\linewidth}
			\centering
			\includegraphics[width=0.85\linewidth]{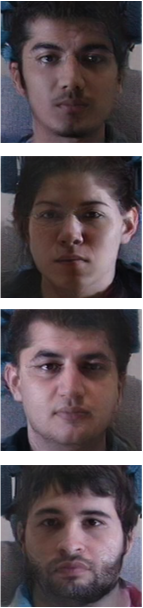}
	\end{minipage}}
	\subfigure[Frontal]{
		\begin{minipage}[b]{0.12\linewidth}
			%\centering
			\includegraphics[width=0.85\linewidth]{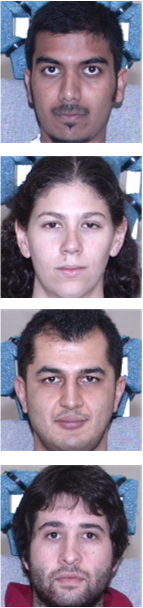}
	\end{minipage}}
	\caption{Face frontalization comparison on the Multi-PIE dataset under the pose of $ 90^\circ $ (first two rows) and $ 75^\circ $ (last two rows).}
	\label{fig:multipie} 
\end{figure}

Fig. \ref{fig:multipie} illustrates the comparison with the state-of-the-art face frontalization methods \cite{hu2018pose, huang2017beyond, yin2017towards,qian2019unsupervised} on the Multi-PIE dataset. In the large pose cases, existing methods are disable to preserve the illumination of profiles on the synthesized results. Face shape and other face components (\eg, eyebrows, mustache and nose) also occur deformation. The reason is those methods are less able to preserve reliable details from the profiles. Compared with the existing methods, our method produces more identity preserving results while keeping the facial details of the profiles as much as possible. In particular, under large poses, our model can recover photo-realistic illumination conditions of the profiles, which is important when frontalized images are used for some other face-related tasks, such as face editing, face pose transfer and face-to-face synthesis. % As for these tasks, the illumination preserving frontalization results generally are more preferred, and it is unusual that the illumination will alter along with the change of face pose.

\begin{figure}[t]
	\centering
	\subfigure[Profile]{
		\begin{minipage}[b]{0.12\linewidth}
			\centering
			\includegraphics[width=0.85\linewidth]{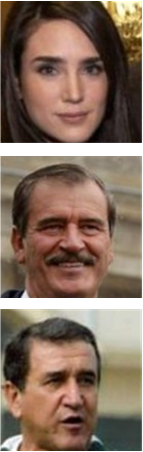}
	\end{minipage}}
	\subfigure[Ours]{
		\begin{minipage}[b]{0.12\linewidth}
			\centering
			\includegraphics[width=0.85\linewidth]{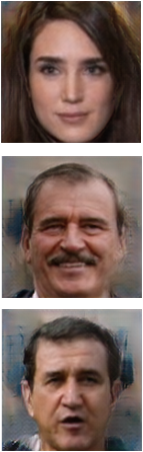}
	\end{minipage}}
	\subfigure[~\cite{zhao2018towards}]{
		\begin{minipage}[b]{0.12\linewidth}
			\centering
			\includegraphics[width=0.85\linewidth]{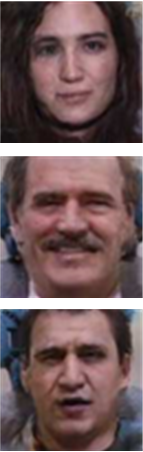}
	\end{minipage}}
	\subfigure[~\cite{huang2017beyond}]{
		\begin{minipage}[b]{0.12\linewidth}
			\centering
			\includegraphics[width=0.85\linewidth]{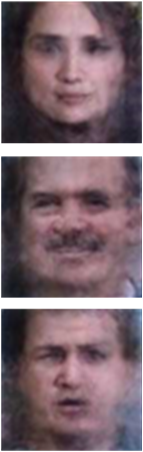}
	\end{minipage}}
	\subfigure[~\cite{tran2017disentangled}]{
		\begin{minipage}[b]{0.12\linewidth}
			\centering
			\includegraphics[width=0.85\linewidth]{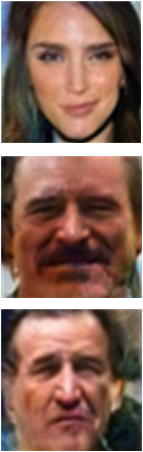}
	\end{minipage}}
	\subfigure[\cite{hassner2015effective}]{
		\begin{minipage}[b]{0.12\linewidth}
			\centering
			\includegraphics[width=0.85\linewidth]{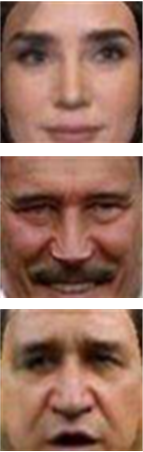}
	\end{minipage}}
	\subfigure[\cite{qian2019unsupervised}]{
	\begin{minipage}[b]{0.12\linewidth}
			\centering
			\includegraphics[width=0.85\linewidth]{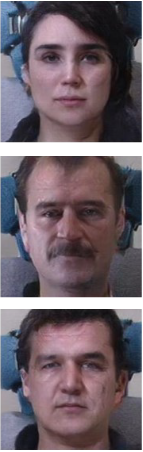}
	\end{minipage}}
	\caption{Face frontalization comparison on the LFW dataset. Our method is trained on Mulit-PIE and tested on LFW.}
	\label{fig:lfw} 
\end{figure}

\begin{table}[t]
	\begin{center}
		\caption{Rank-1 recognition rates (\%) across poses under Setting 2 of the Multi-PIE. The best two results are highlighted by \textbf{bold} and \underline{underline} respectively.}
		\label{table:Multipie-Setting2-FRR}
		\begin{tabular}{lccccccc}
			\hline\noalign{\smallskip}
			Method& \(\pm 15^{\circ}\) & \(\pm 30^{\circ}\) & \(\pm 45^{\circ}\) & \(\pm 60^{\circ}\) & \(\pm 75^{\circ}\) & \(\pm 90^{\circ}\) & Avg \\
			\hline
			Light CNN \cite{wu2018light}& 98.59 & 97.38 & 92.13 & 62.09 & 24.18 & 5.51 & 63.31 \\
			DR-GAN \cite{tran2017disentangled}& 94.90 & 91.10 & 87.20 & 84.60 & - & - & 89.45 \\
			FF-GAN \cite{yin2017towards}& 94.60 & 92.50 & 89.70 & 85.20 & 77.20 & 61.20 & 83.40 \\
			TP-GAN \cite{huang2017beyond}& 98.68 & 98.06 & 95.38 & 87.72 & 77.43 & 64.64 & 86.99 \\
			CAPG-GAN \cite{hu2018pose} & 99.82 & 99.56 & 97.33 & 90.63 & 83.05 & 66.05 & 89.41 \\
			PIM \cite{zhao2018towards} & 99.30 & 99.00 & 98.50 & 98.10 & 95.00 & 86.50 & 96.07 \\
			3D-PIM \cite{zhao20183d} & 99.64 & 99.48 & 98.81 & 98.37 & 95.21 & 86.73 & 96.37 \\
			DA-GAN \cite{yin2020dual} & \underline{99.98} & \underline{99.88} & 99.15 & 97.27 & 93.24 & 81.56 & 95.18 \\
			HF-PIM \cite{cao2018learning} & \textbf{99.99} & \textbf{99.98} & \textbf{99.98} & \textbf{99.14} & \underline{96.40} & \underline{92.32} & \underline{97.97} \\ %97.97
			\hline
			\textbf{Ours} & 99.86 & 99.80 & \underline{99.37} & \underline{98.85} & \textbf{97.20} & \textbf{93.17} & \textbf{98.04} \\ % 98.04
			\hline
		\end{tabular}
	\end{center}
\end{table}

We further qualitatively compare face frontalization results of our model on the LFW dataset with \cite{zhao2018towards, huang2017beyond, tran2017disentangled, hassner2015effective,qian2019unsupervised}. As shown in Fig.~\ref{fig:lfw}, the existing methods fail to recover clear global structures and fine facial details. Also they cannot preserve the illumination of the profiles. Though FNM\cite{qian2019unsupervised} generates high qualitative images, it is still disable to preserve identity. It is worth noting that our method produces more photo-realistic faces with identity and illumination well-preserved, which also demonstrates the generalizability of our model in the uncontrolled environment. More results under large poses are provided in the supplementary material.

\subsection{Quantitative evaluation}

In this subsection, we quantitatively compare the proposed method with other methods in terms of recognition accuracy on Multi-PIE and LFW. The recognition accuracy is calculated by firstly extracting deep features with LightCNN-29\cite{wu2018light} and then measuring similarity between features with a cosine-distance metric.

Tab.~\ref{table:Multipie-Setting2-FRR} shows the Rank-1 recognition rates of different methods under Setting 2 of Multi-PIE. Our method has advantages over competitors, especially at large poses (\eg, $ 75^\circ $, $ 90^\circ $), which demonstrates that our model can synthesize frontal images while preserving the identity information. The recognition rates under Setting 1 is provided in the supplementary material.

\begin{table}[t]
	\begin{center}
		\caption{Face verification accuracy (ACC) and area-under-curve (AUC) results on LFW.}
		\label{table:LFW-ACC-AUC}
		\begin{tabular}{lcccccc}
			\hline\noalign{\smallskip}
			Method & FaceNet\cite{schroff2015facenet} & VGG Face\cite{parkhi2015deep} & FF-GAN\cite{yin2017towards} & CAPG-GAN\cite{hu2018pose} & DA-GAN\cite{yin2020dual} & \textbf{Ours} \\
			\hline
			ACC(\%) & 99.63 & 98.95 & 96.42 & 99.37 & 99.56	& \textbf{99.65} \\
			AUC(\%) & - & -  & 99.45 & 99.90 & 99.91 & \textbf{99.92} \\
			\hline
		\end{tabular}
	\end{center}
\end{table}

Tab.~\ref{table:LFW-ACC-AUC} compares the face verification performance (ACC and AUC) of our method with other state-of-the-arts\cite{schroff2015facenet, parkhi2015deep, yin2017towards, hu2018pose, yin2020dual} on the LFW. Our method achieves 99.65 on accuracy and 99.92 on AUC, which is also comparable with other state-of-the-art methods. The above quantitative results prove that our method is able to preserve the identity information effectively.

\subsection{Ablation Study}

\begin{figure}[t]
	\centering
	\includegraphics[height=4.0cm]{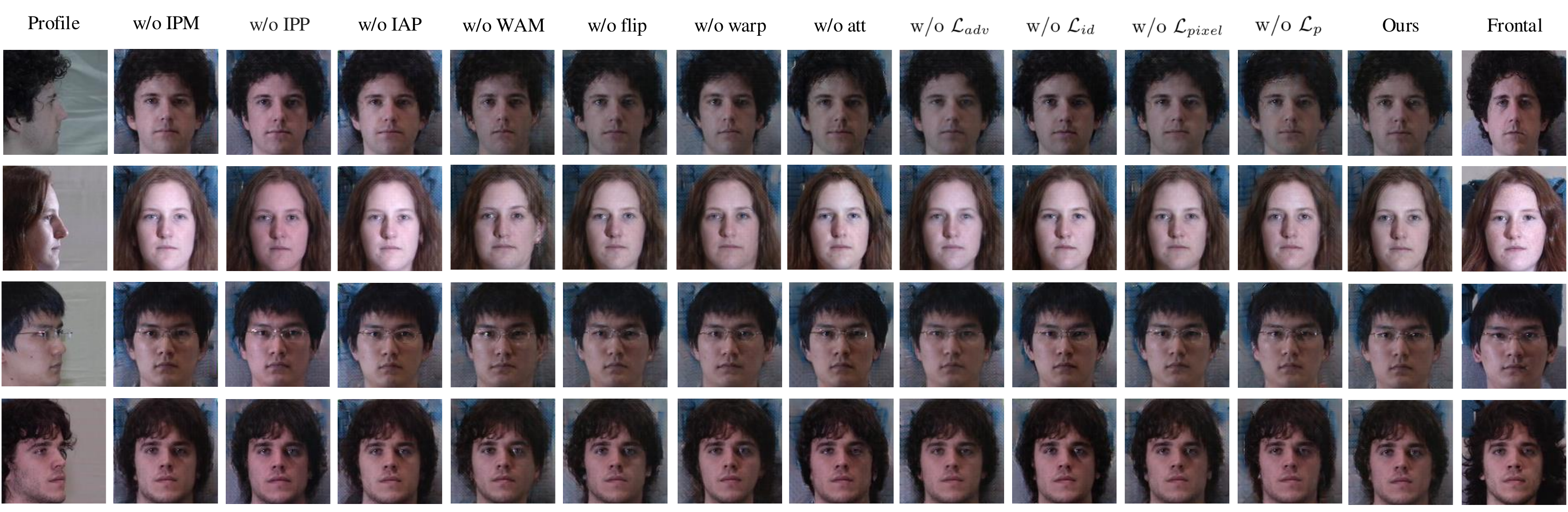}
	\caption{Model Comparsion: synthesis results of our model and its variants on Multi-PIE}
	\label{fig:ablationall.pdf}
\end{figure}

\begin{table}[t]
	\begin{center}
		\caption{Incomplete variants analysis: Rank-1 recognition rates (\%) across poses under Setting 2 of the Multi-PIE dataset. IAP and IPP denote the illumination adaption pathway and illumination preserving pathway in the Illumination Preserving Module (IPM). Warp, flip and att denote the three variants in Warp Attention Module (WAM).}
		\label{table:Multipie-Setting2-Ablation}
		\begin{tabular}{lccccccc}
			\hline\noalign{\smallskip}
			Method& \(\pm 15^{\circ}\) & \(\pm 30^{\circ}\) & \(\pm 45^{\circ}\) & \(\pm 60^{\circ}\) & \(\pm 75^{\circ}\) & \(\pm 90^{\circ}\)& Avg \\
			\hline
			
			w/o IPM & 99.83 & 99.77 & 99.35 & 98.74 & 97.18 & 93.15 &  98.00 \\
			IPM w/o IPP & 99.84 & 99.74 & 99.36 & 98.47 & 96.73 & 91.56 & 97.62 \\
			IPM w/o IAP & 99.83 & 99.76 & 99.30 & 98.70 & 97.11 & 92.83 & 97.92 \\
			\hline
			w/o WAM & 99.84 & 99.46 & 98.91 & 97.27 & 93.18 & 86.23 & 95.81 \\			
			WAM w/o flip  & 99.84 & 99.69 & 99.27 & 98.10 & 96.57 & 92.65 & 97.69 \\
			WAM w/o warp  & 99.83 & 99.64 & 99.16 & 97.83 & 94.60 & 88.16 & 96.54 \\
			WAM w/o att & 99.85 & 99.79 & 99.36 & 98.71 & 96.81 & 93.09 & 97.94 \\
			
			\hline
			w/o $ \mathcal{L}_{adv} $ & 99.83 & 99.72 & 99.28 & 98.57 & 97.09 & 93.11 & 97.93 \\
			w/o $ \mathcal{L}_{id} $ & 99.85 & 99.62 & 99.12 & 97.42 & 93.93 & 86.05 & 96.00 \\
			w/o $ \mathcal{L}_{pixel} $ & 99.83 & 99.77 & 99.35 & 98.79 & 97.05 & 92.85 & 97.94 \\
			
			w/o $ \mathcal{L}_{p} $ & 99.81 & 99.75 & 99.33 & 98.62 & 97.13 & 93.10 & 97.96\\
			\hline
			\textbf{Ours} & \textbf{99.86} & \textbf{99.80} & \textbf{99.37} & \textbf{98.85} & \textbf{97.20} & \textbf{93.17} & \textbf{98.04}\\
			\hline
		\end{tabular}
	\end{center}
\end{table}

In this subsection, we analyze the respective roles of the different modules and loss functions in frontal view synthesis. Both qualitative perceptual performance (Fig.~\ref{fig:ablationall.pdf}) and face recognition rates (Tab.~\ref{table:Multipie-Setting2-Ablation}) are reported for comprehensive comparison under the Multi-PIE Setting 2. We can see that our FFWM exceeds all its variants in both quantitative and qualitative evaluations.

\subsubsection{Effects of the Illumination Preserving Module (IPM).}
Although without IPM the recognition rates drop slightly (as shown in Tab.~\ref{table:Multipie-Setting2-Ablation}), the synthesized results cannot preserve illumination and are approximate to the inconsistent ground-truth illumination (as shown in Fig.~\ref{fig:ablationall.pdf}). We also explore the contributions of illumination adaption pathway (IAP) and illumination preserving pathway (IPP) in the IPM. As shown in Fig.~\ref{fig:ablationall.pdf}, without IPP, the illumination of synthesized images tend to be inconsistent with the profiles and ground-truth images. And without IAP, the illumination of synthesized images tends to be a tradeoff between the profiles and the illumination inconsistent ground-truth images. Only integrating IPP and IAP together, our model can achieve illumination preserving image synthesis. Furthermore, our model archives a lower recognition rate when removing the IPP, which demonstrates that the IPP prompts the synthesized results to keep reliable information of the profiles.

\subsubsection{Effects of the Warp Attention Module (WAM).} 
We can see that without WAM, the synthesized results tend to be smooth and distorted in the self-occlusion parts (as shown in Fig.~\ref{fig:ablationall.pdf}). As shown in Tab.~\ref{table:Multipie-Setting2-Ablation}, without WAM, the recognition rates drop significantly, which proves that WAM dominates in preserving identity information. Moreover, we explore the contributions of three components in the WAM, including taking flipped warped feature as additional input (w/o flip), feature warping (w/o warp) and feature attention (w/o att). As shown in Fig.~\ref{fig:ablationall.pdf}, taking flip feature as additional input has benefits on recovering the self-occlusion parts on the synthesized images. Without the feature attention mechanism, there are artifacts on the synthesized images. Without feature warping, the synthesized results get worse visual performance. These results above suggest that each component in WAM is essential for synthesizing identity preserving and photo-realistic frontal images.

\subsubsection{Effects of the losses.} 
As shown in Tab.~\ref{table:Multipie-Setting2-Ablation}, the recognition rates decrease if one loss function is removed. Particularly, the rates drop significantly for all poses if the $ \mathcal{L}_{id} $ loss is not adapted. We also report the qualitative visualization results in Fig.~\ref{fig:ablationall.pdf}. Without $ \mathcal{L}_{adv} $ loss, the synthesized images tend to be blurry, suggesting the usage of adversarial learning.  Without $ \mathcal{L}_{id} $ and $ \mathcal{L}_{pixel} $, our model cannot promise the visual performance on the local textures (\eg, eyes). Without $\mathcal{L}_p$, the synthesized faces present artifacts at the edges (\eg, face and hair).

\section{Conclusion}

In this paper, we propose a novel Flow-based Feature Warping Model (FFWM) to effectively address the challenging problem in face frontalization, photo-realistic and illumination preserving image synthesis with illumination inconsistent supervision. Specifically, an Illumination Preserve Module is proposed to address the illumination inconsistent issue. It helps FFWM to synthesize photo-realistic frontal images while preserving the illumination of profile images. Furthermore, the proposed Warp Attention Module reduces the pose discrepancy in the feature space and helps to synthesize frontal images effectively.  Experimental results demonstrate that our method not only synthesizes photo-realistic and illumination preserving results but also outperforms state-of-the-art methods on face recognition across large poses.

\subsection*{Acknowledgement}

This work is partially supported by the National Natural Science Foundation of China (NSFC) under Grant No.s 61671182 and U19A2073.

\bibliographystyle{splncs04}
\bibliography{egbib}

\clearpage

\section*{Appendix}

\subsection*{A. \quad Training details for Flow Estimation Networks}

It is difficult and expensive to manually obtain two ground-truth flow fields between the profile $ I $ and the ground-truth image $ I^{gt} $. Instead, we introduce the landmark loss\cite{li2018learning}, sampling correctness loss\cite{ren2020deep} and the regularization term \cite{ren2020deep} to pretrain the bi-directional flow estimation networks (the forward flow estimation network $ \mathcal{F} $ and the reverse flow estimation network $ \mathcal{F'} $). For landmark  loss, we use the dense landmark detection method\footnote{https://www.faceplusplus.com/dense-facial-landmarks/} to detect 1000 facial landmarks for $ I $ and $ I^{gt} $. We then move face contour landmarks in the vertical and horizontal directions and mark new landmarks and correspondences in the above areas between $ I $ and $ I^{gt} $. In this way, we can deform the additional face areas (\eg, hair, neck and ears). In our experiments, we pretrain the $ \mathcal{F} $ and the $ \mathcal{F'} $ for 4 epochs and then all networks are trained in an end-to-end manner.

\subsection*{B. \quad Additional Qualitative Results}

\begin{figure}[t]
	\centering
	\includegraphics[width=0.95\textwidth]{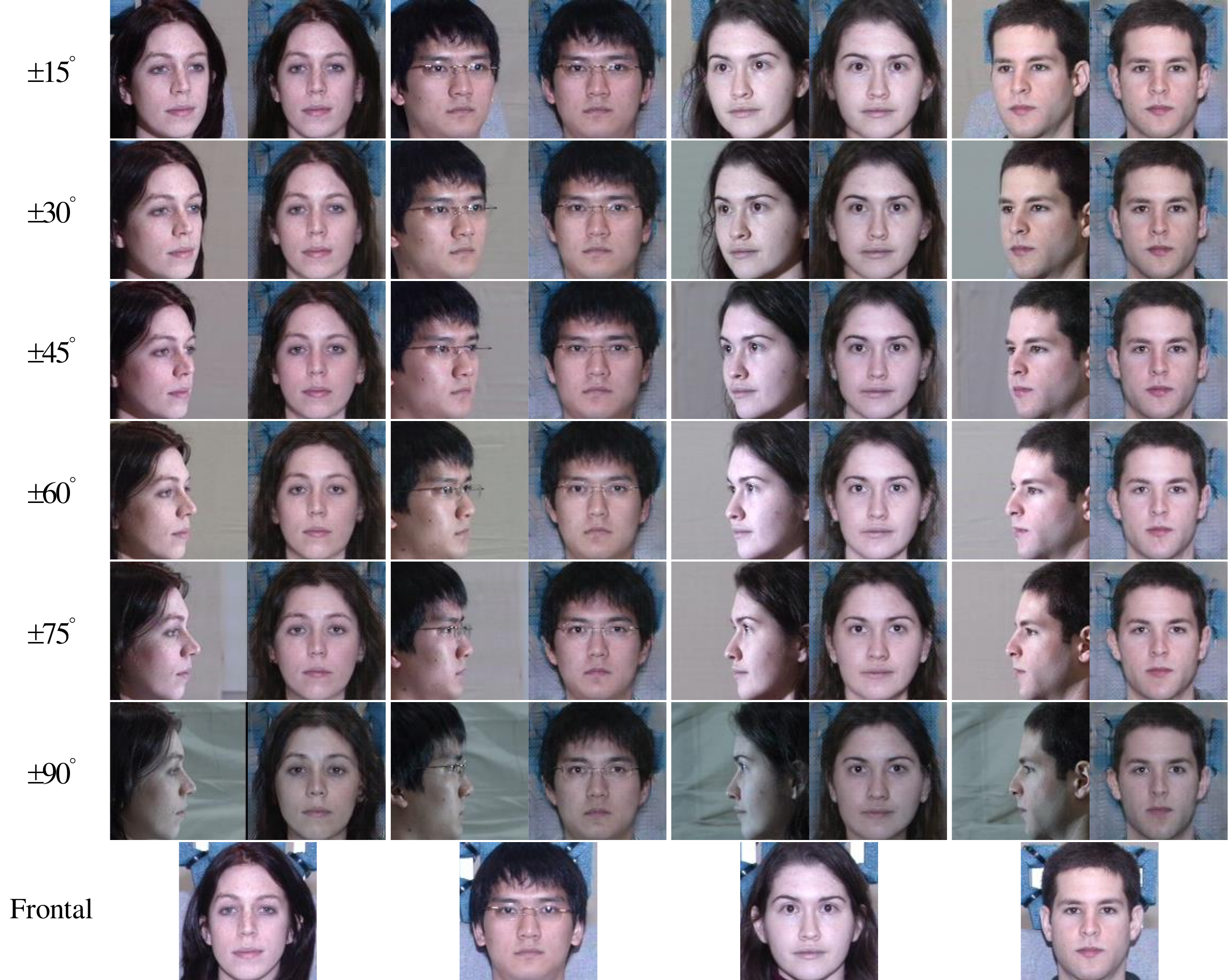}
	\caption{Synthesis results by our model under different poses of the Multi-PIE dataset. From top to down, the poses are $ \pm15^\circ, \pm30^\circ, \pm45^\circ , \pm60^\circ, \pm75^\circ, \pm90^\circ$. The ground-truth frontal images are provided at the last row.}
	\label{fig:result}
\end{figure}

Fig.~\ref{fig:result} shows the face synthesized results on Multi-PIE within $ \pm90^\circ $  at 12 different poses (except $ 0^\circ $). It is obvious that our model can synthesize photo-realistic images with delicate details across all pose variations.

\begin{figure}[t]
	\centering
	\includegraphics[width=0.95\textwidth]{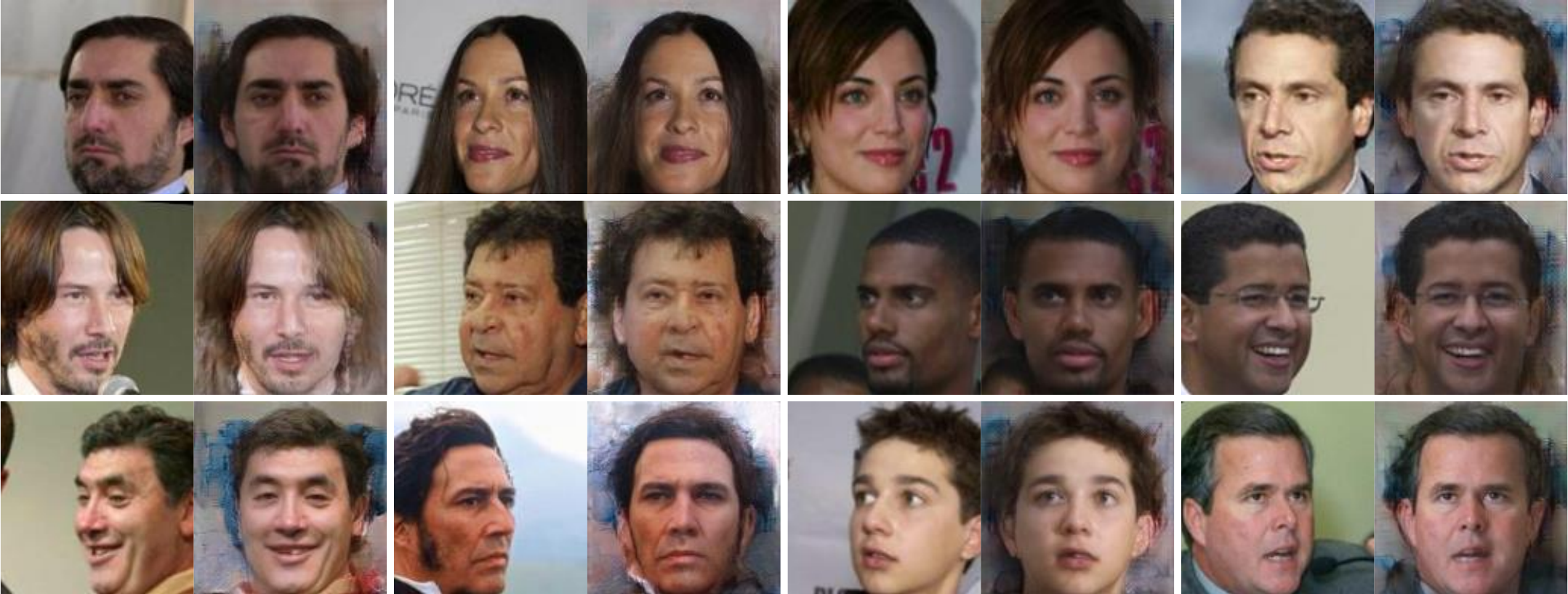}
	\caption{Additional synthesis results of the LFW dataset by our model. Each pair presents the profile (left) and the synthesized frontal face (right).}
	\label{fig:lfw_largepose}
\end{figure}

More synthesized results on the LFW dataset under large poses are given in Fig.~\ref{fig:lfw_largepose}. It can be seen that our method exhibits satisfying generalization to in-the-wild face images, and the frontalization results are consistent with the profile face images.

\begin{figure}[t]
	\centering
	\includegraphics[width=0.3\textwidth]{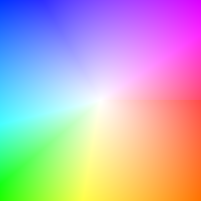}
	\caption{Flow field color coding used in this paper. The displacement of every pixel in this illustration is the vector from the center of the square to this pixel. The central pixel does not move.}
	\label{fig:flow_color}
\end{figure}

\begin{figure}[t]
	\centering
	\includegraphics[width=0.95\textwidth]{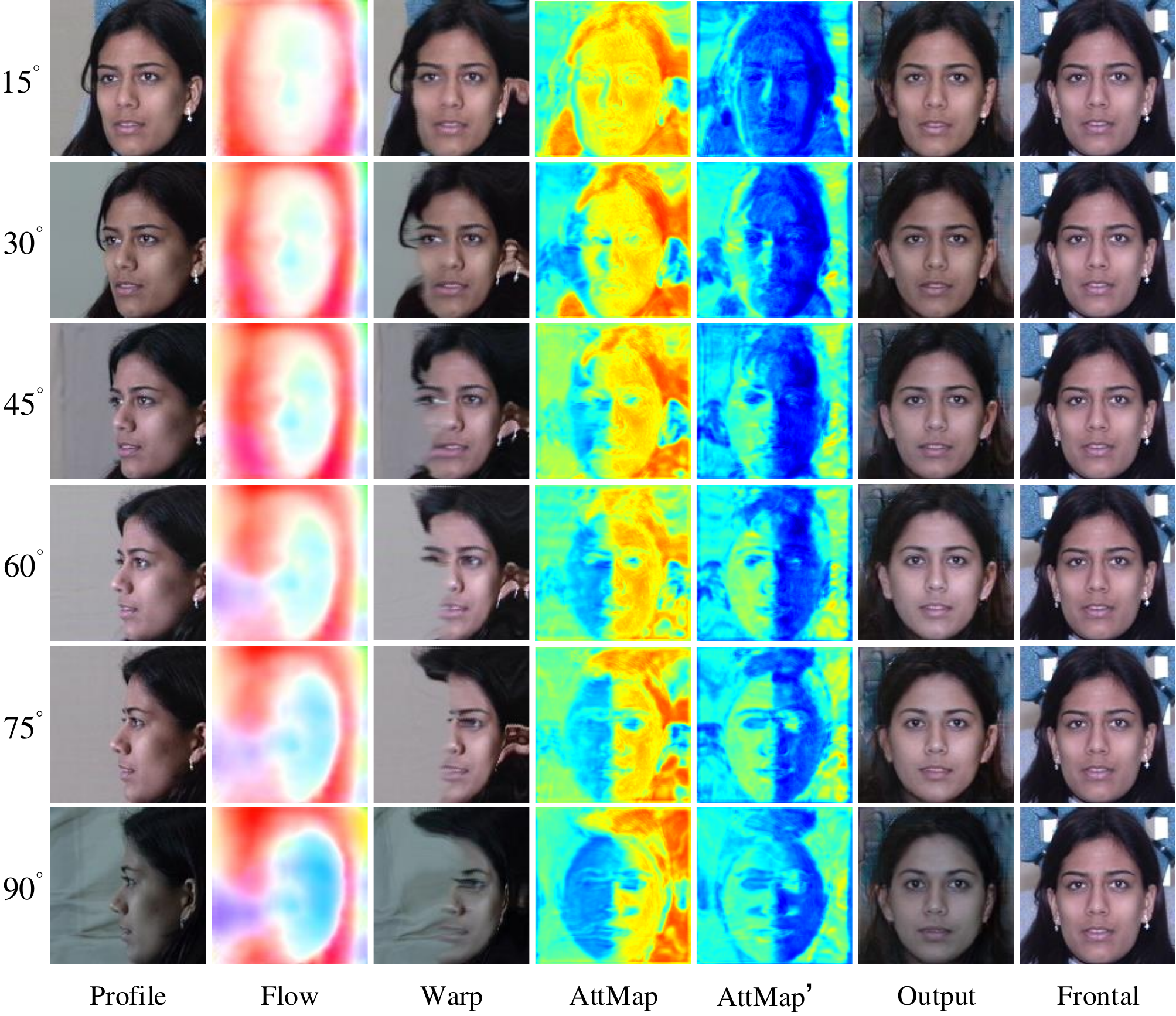}
	\caption{The visualization of learned flow fields, warped images, and attention maps. AttMap and AttMap' represent the attention map of the warped feature and its flip, respectively}
	\label{fig:attention_flow}
\end{figure}

To better understand the Warp Attention Module (WAM), we visualize the learned flow fields, warped images, and attention maps in the Fig.~\ref{fig:attention_flow}. For optical flow visualization, we use the color coding of Butler \etal \cite{butler2012naturalistic}. The color coding scheme is illustrated in Fig.~\ref{fig:flow_color}. Hue represents the direction of the displacement vector, while the intensity of the color represents its magnitude. White color corresponds to no motion. As shown in Fig.~\ref{fig:attention_flow}, face frontalization can be viewed as the horizontal rotation of the face (the learned flow fields are mainly blue and red which represent horizontal rotation). Using learned forward flows field can warp the profile to the frontal view. And the learned attention maps can help focus on the critical parts of the warped features.

\subsection*{C. \quad Additional Quantitative Results}

Tab.~\ref{table:Multipie-Setting1-FRR} shows the Rank-1 recognition rates of different methods under Multi-PIE Setting 1. The recognition rates of all methods drop as pose degree increases. Missing more facial appearance information leads to the difficulty of synthesis task with pose rotation angle increases. As shown in Tab.~\ref{table:Multipie-Setting1-FRR}, our model achieves the best performance across all poses, which demonstrates that our model can synthesize frontal images while preserving the identity information.

\begin{table}[t]
	\begin{center}
		\caption{Rank-1 recognition rates (\%) across poses under Setting 1 of the Multi-PIE. The best two results are highlighted by \textbf{bold} and \underline{underline} respectively.}
		\label{table:Multipie-Setting1-FRR}
		\begin{tabular}{lccccccc}
			\hline\noalign{\smallskip}
			Method& \(\pm 15^{\circ}\) & \(\pm 30^{\circ}\) & \(\pm 45^{\circ}\) & \(\pm 60^{\circ}\) & \(\pm 75^{\circ}\) & \(\pm 90^{\circ}\) & Avg \\
			\hline
			Light CNN \cite{wu2018light}& 99.78 & 99.80 & 97.45 & 73.30 & 32.35 & 9.00 & 68.61 \\
			TP-GAN \cite{huang2017beyond}& 99.78 & \underline{99.85} & 98.58 & 92.93 & 84.10 & 64.03 & 89.88 \\
			CAPG-GAN \cite{hu2018pose} & \underline{99.95} & 99.37 & 98.28 & 93.74 & 87.40 & \underline{77.10} & 92.64 \\
			PIM \cite{zhao2018towards} & 99.80 & 99.40 & 98.30 & 97.70 & 91.20 & 75.00 & 93.57 \\
			3D-PIM \cite{zhao20183d} & 99.83 & 99.47 & \underline{99.34} & \underline{98.84} & \underline{94.34} & 76.12 & \underline{94.66} \\
			FNM \cite{qian2019unsupervised} & 99.90 & 99.50 & 98.20 & 93.70 & 81.30 & 55.80 & 88.07 \\
			\hline
			\textbf{Ours} & \textbf{100.00} & \textbf{100.00} & \textbf{100.00} & \textbf{98.86} & \textbf{96.54} & \textbf{88.55} & \textbf{97.33} \\
			\hline
		\end{tabular}
	\end{center}
\end{table}

\end{document}